\documentclass[twocolumn]{svjour3}         
\smartqed  
\usepackage{graphicx}

\usepackage{titlesec} 
\titleformat{\paragraph}[runin]
{\bfseries\scshape}{\theparagraph}{1em}{}
\titlespacing{\paragraph}{0em}{1ex}{.5em} 
\usepackage{todonotes}
\usepackage{placeins} 
\usepackage{booktabs}
\usepackage{tabularx}
\usepackage{multirow}
\usepackage{multicol}
\usepackage{bm}
\usepackage{hyperref}
\usepackage{url}
\usepackage{balance}
\usepackage{enumitem}
\usepackage{color}
\usepackage{bbding}
\usepackage{nccmath}
\usepackage{amsmath,amssymb}
\usepackage{graphicx}
\usepackage{cite}
\usepackage{makecell}
\usepackage{subfig}
\usepackage{graphicx}
\usepackage{amsmath}
\usepackage{amssymb}
\usepackage{color}
\usepackage{epstopdf}
\usepackage{array}
\usepackage{multirow}
\usepackage{amsfonts}
\usepackage{booktabs}
\usepackage{float}
\usepackage{algorithm}
\usepackage{algpseudocode}
\usepackage[switch]{lineno}

\usepackage{xcolor}
\usepackage{multirow}
\usepackage{booktabs}

\definecolor{betterred}{RGB}{200, 0, 0}   
\definecolor{worsegreen}{RGB}{0, 150, 0} 

\newcommand{\gain}[1]{\textcolor{betterred}{\textbf{#1}}}  
\newcommand{\loss}[1]{\textcolor{worsegreen}{#1}}          

\begin{document}

\title{Geometry-Editable and Appearance-Preserving Object Composition}

\author{Jianman Lin \and
        Haojie Li \and
        Chunmei Qing \and
        Zhijing Yang \and
        Liang Lin \and
        Tianshui Chen
}

\institute{Jianman Lin \at
             South China University of Technology, Guangzhou, China \\
              \email{linjianmancjx@gmail.com} 
           \and
           Haojie Li \at
             South China University of Technology, Guangzhou, China \\
              \email{12hjli4@gmail.com}
           \and
           Chunmei Qing \at
             South China University of Technology, Guangzhou, China \\
              \email{qchm@scut.edu.cn} 
           \and
           Zhijing Yang \at
             Guangdong University of Technology, Guangzhou, China \\
              \email{yzhj@gdut.edu.cn} 
           \and
           Liang Lin \at
             Sun Yat-sen University, Guangzhou, China \\
              \email{linliang@ieee.org}
           \and
           Tianshui Chen \at
             Guangdong University of Technology, Guangzhou, China \\
              \email{tianshuichen@gmail.com}
}

\date{Received: date / Accepted: date}

\maketitle


\def\eg{\emph{e.g.}}
\def\etal{\emph{et al}}

\newcommand{\EQREF}{Eq.~\eqref}
\newcommand{\EQSREF}{Eqs.~\eqref}
\newcommand{\FIGREF}{Fig.~\ref}
\def\proposed{VB} 
\def\fixcolor{black}
\def\hstate{\bm {\tilde s}}
\def\rstate{\bm s}
\def\jstate{\bm s^{jn}}
\def\jpolicy{\overrightarrow{\pi}}
\def\vpref{v_{\text {pref}}}
\def\vector#1{\mbox{\boldmath $#1$}}
\def\sup#1{^{(\rm #1)}}
\def\sub#1{_{\rm #1}}
\def\supi#1{^{(#1)}}
\def\vct#1{\mbox{\boldmath $#1$}}
\def\eg{{\it e.g.}}
\def\cf{{\it c.f.}}
\def\ie{{\it i.e.}}
\def\etal{{\it et al. }}
\def\etc{{\it etc}}
\newcommand{\argmax}{\mathop{\rm argmax}\limits}
\newcommand{\argmin}{\mathop{\rm argmin}\limits}

\def\Rerr{\Delta \bm r}
\def\Terr{\Delta \bm t}
\def\Xerr{\Delta \bm x}
\def\XerrRel{\Delta \bm {\tilde x}}
\def\Xgt{\dot{\bm x}}
\def\Rgt{\dot{R}}
\def\Tgt{\dot{\bm t}}
\def\arraystretchlen{1.0}

\def\cam{c}
\def\image{\mathcal I}
\def\traj{\mathcal X}
\def\btraj{\mathcal {\bm X}}
\def\keypoints{\mathcal P}
\def\states{\mathcal S}
\def\bstates{\mathcal {\bm S}}
\def\state{\bm s}
\def\ped{\bm x}
\def\pedi{\bm p} 
\def\obs{\bm z}

\def\Fi{\bm F_r}
\def\Fp{\bm F_p}
\def\vpref{\bm w}
\def\ENERGY{{\mathcal E}}

\def\DIFF#1{\textcolor{black}{#1}}
\def\DIFFCR#1{\textcolor{black}{#1}}

\begin{abstract}
General object composition (GOC) aims to seamlessly integrate a target object into a background scene with desired geometric properties, while simultaneously preserving its fine-grained appearance details. Recent approaches derive semantic embeddings and integrate them into advanced diffusion models to enable geometry-editable generation. However, these highly compact embeddings encode only high-level semantic cues and inevitably discard fine-grained appearance details. We introduce a Disentangled Geometry-editable and Appearance-preserving Diffusion (DGAD) model that first leverages semantic embeddings to implicitly capture the desired geometric transformations and then employs a cross-attention retrieval mechanism to align fine-grained appearance features with the geometry-edited representation, facilitating both precise geometry editing and faithful appearance preservation in object composition. Specifically, DGAD builds on CLIP/DINO-derived and reference networks to extract semantic embeddings and appearance-preserving representations, which are then seamlessly integrated into the encoding and decoding pipelines in a disentangled manner. We first integrate the semantic embeddings into pre-trained diffusion models that exhibit strong spatial perception capabilities to implicitly capture object geometry, thereby facilitating flexible object manipulation and ensuring effective editability. Then, we design a dense cross-attention mechanism that leverages the implicitly learned object geometry to retrieve and spatially align appearance features with their corresponding regions, ensuring faithful appearance consistency. Extensive experiments on public benchmarks demonstrate the effectiveness of the proposed DGAD framework.

\end{abstract}

\begin{figure*}[htp]
  \centering
  \includegraphics[width=0.95\textwidth]{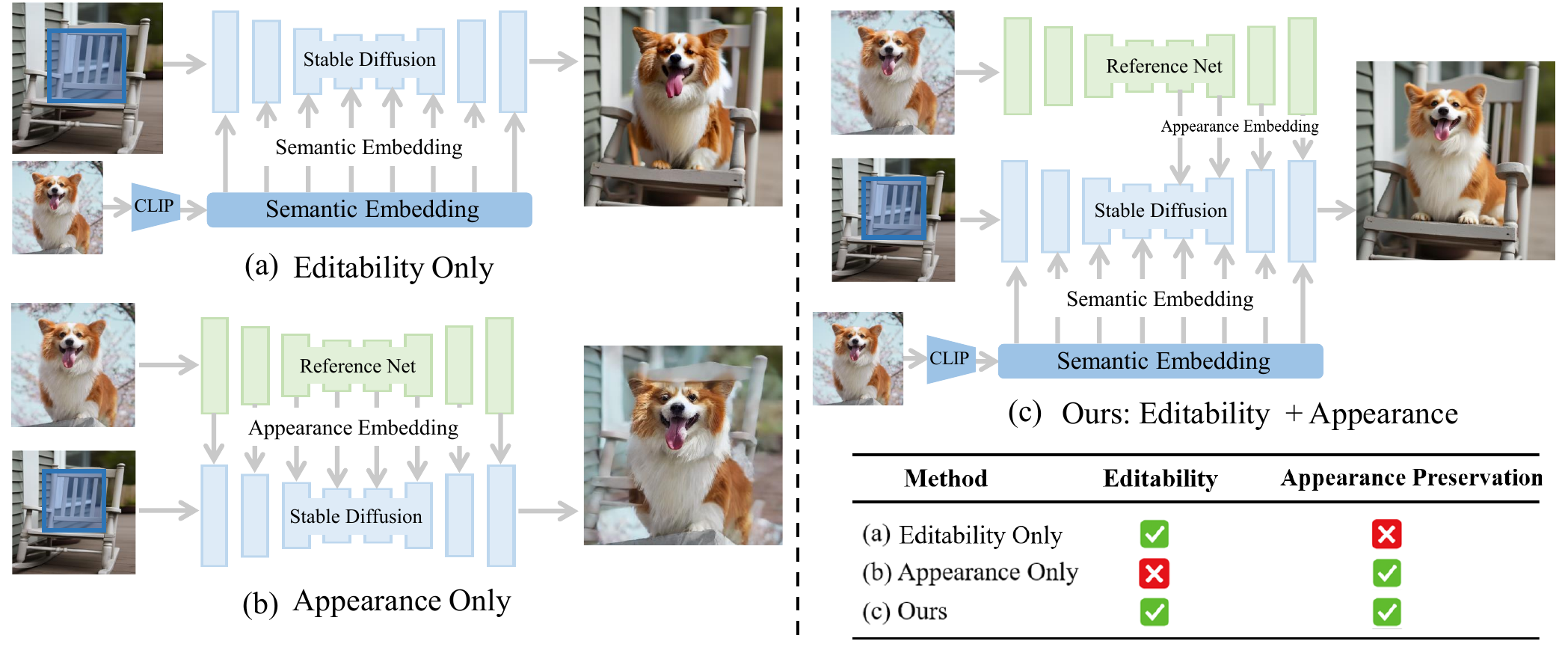} 
  \caption{(a) Leverages compact semantic embeddings to enable object editability but fails to preserve appearance details.
(b) Utilizes appearance features to retain visual fidelity, yet lacks editing capability.
Unlike both, Our method implicitly learns the geometry-editable representation and explicitly aligns fine-grained appearance features with the geometry-edited representation, facilitating both precise geometry editing and faithful appearance preservation.}
\label{fig: motivation}
\end{figure*}

\section{Introduction}
General Object composition (GOC) involves editing a target object to seamlessly integrate it into a background scene at arbitrary angles and positions, while preserving the object’s appearance details without alteration. A robust generative object composition system should automate complex tasks such as interactive image editing, virtual environment creation, and content generation for augmented and virtual reality (AR/VR) applications, by simultaneously supporting flexible object editing and faithful appearance preservation — a critical balance that minimizes the need for manual adjustment in both geometric editing and visual consistency. Thanks to the emergence of large-scale pre-trained diffusion models \cite{nichol2021improved, song2020denoising, ho2022classifier, Hu2023LaMDLM}, substantial progress has been made in generative compositing. However, these models still face challenges in simultaneously editing the target object according to the user's expectations and maintaining consistent appearance details, limiting their applicability in real-world scenarios.

Existing methods either introduce CLIP/DINO derived semantic embeddings \cite{song2023objectstitch, chen2024anydoor, yang2023paint}, or leverage pixel-aligned appearance features to achieve object composition \cite{zhao2024zero}, as shown in Fig. \ref{fig: motivation}. However, both approaches fail to simultaneously ensure object editability and appearance preservation. The former encodes the object into a compact semantic embedding, which demonstrates strong compatibility with pre-trained diffusion models and enables robust manipulation of geometric properties such as shape deformation and viewpoint changes. Yet, this compact encoding inevitably loses high-frequency details, making it difficult to preserve appearance during editing. In contrast, the latter employs reference networks to extract pixel-wise appearance features, which maintain tight spatial correspondence with the object and excel at reconstructing appearance details from noise latents. Nevertheless, the rigid spatial alignment severely restricts editing flexibility, often resulting in copy-paste-like outputs rather than adaptive transformations.  Object composition requires both object editability and appearance preservation, and designing an effective mechanism to leverage the advantages of both is key to solving the problem.

To address these challenges, we first leverage semantic embeddings to implicitly capture the desired geometric transformations, and then employ a dense cross-attention retrieval mechanism to align fine-grained appearance features with the geometry-edited representation. 
This approach stands in contrast to prior methods, which often suffer from inherent limitations. 
Specifically, methods like \cite{ye2023ip, yang2023paint} rely on precise object masks to explicitly encode geometric properties, thereby constraining editing flexibility and efficiency. 
Conversely, approaches such as \cite{zhao2024zero} perform implicit retrieval via standard cross-attention, frequently resulting in suboptimal appearance preservation due to spatial misalignment between geometry and appearance. By comparison, we leverage the inherent spatial perception capabilities of pretrained diffusion models to implicitly capture object geometry, subsequently employing position-wise retrieval to strictly align appearance features with their corresponding geometric regions. This design ensures both editability and faithful appearance preservation.

To this end, we propose the Disentangled Geometry-editable and Appearance-preserving Diffusion (DGAD) model, which leverages compact semantic embeddings to implicitly learn the geometric properties of objects during encoding, and employs the resulting representations to explicitly retrieve and spatially map appearance features to corresponding geometric regions during decoding.  Specifically, during the encoding stage, the initial input is constructed by concatenating the user-specified regions with surrounding contextual information. We leverage CLIP/DINO-derived semantic embeddings \cite{radford2021learning, oquab2023dinov2} and a cross-attention mechanism built upon the strong spatial perception capabilities of pretrained diffusion models to implicitly learn the geometric properties of objects, which in turn enables flexible manipulation and thereby ensures object editability. To ensure consistent object appearance, we introduce a dense cross-attention mechanism that leverages encoded features to establish explicit correspondences with appearance features from a reference network. The encoded features, which capture both semantic and geometric properties, serve as queries, while the appearance features act as keys and values. To further guide the attention toward semantically relevant regions, a position-wise gating weight is learned from the query features to explicitly represent the object’s geometric structure, enabling the model to adaptively retrieve and align appearance features with their corresponding  geometric regions. This mechanism is applied exclusively during the decoding stage, where appearance retrieval is conditioned on fully geometry-edited representation, thereby enhancing both object editability and appearance preservation in the composition process.

Our contributions are threefold. First, we propose the Disentangled Geometry-editable and Appearance-preserving Diffusion (DGAD) model, a novel framework that is the first to explicitly disentangle geometry editing from appearance preservation in object composition. It achieves this by implicitly learning geometry during encoding and explicitly retrieving appearance features during decoding. Second, we introduce a dense attention mechanism that establishes position-aware correspondences between the edited geometry and original appearance, ensuring high-fidelity preservation under complex transformations. Finally, extensive experiments demonstrate that DGAD significantly outperforms state-of-the-art methods in both geometry editability and appearance preservation.

\begin{figure*}[htp]
  \centering
  \includegraphics[width=0.95\textwidth]{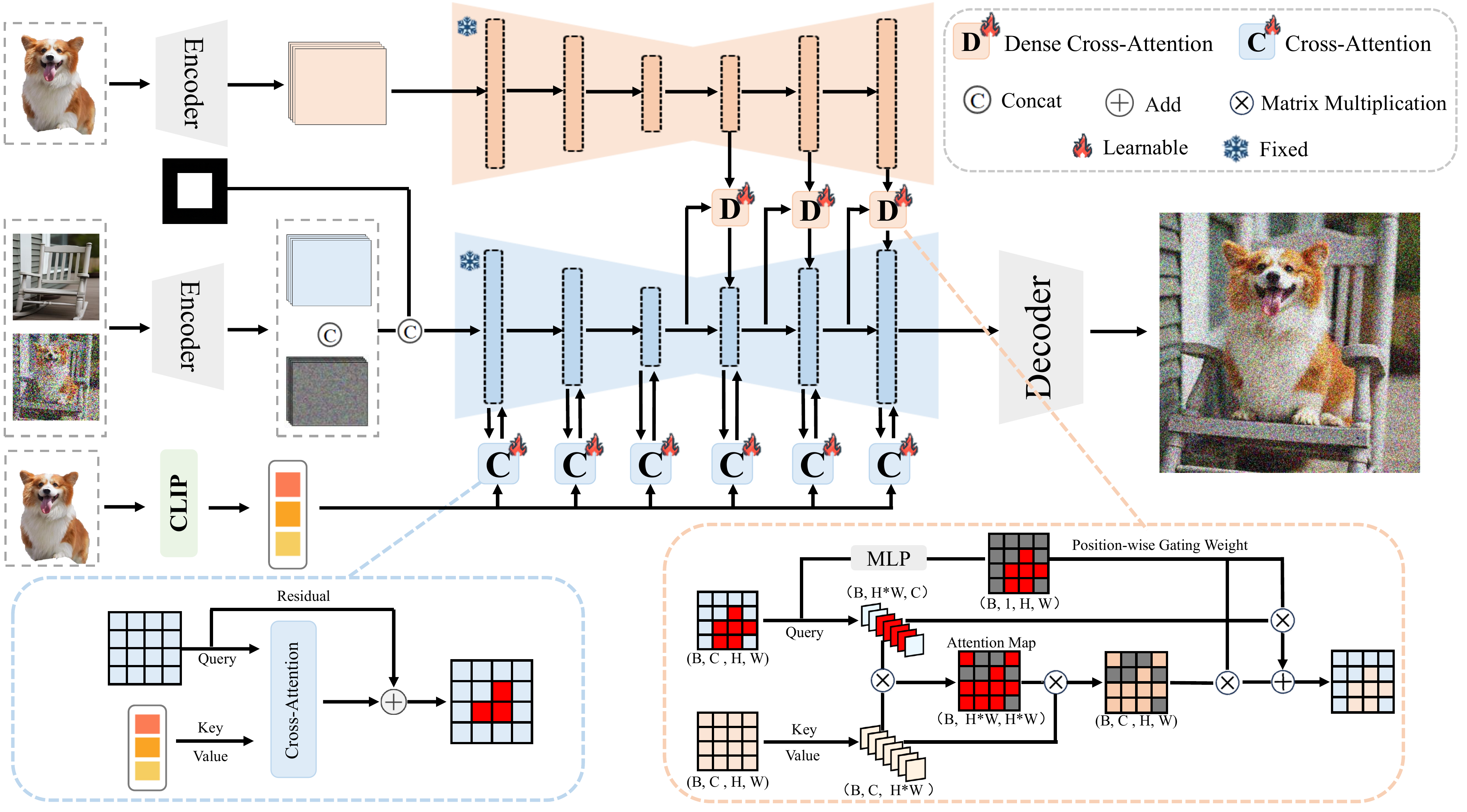} 
  \caption{Overview of the proposed Disentangled Geometry-editable and Appearance-preserving Diffusion (DGAD) framework. The model leverages semantic embeddings to implicitly capture the desired geometric transformations, and subsequently employs a dense cross-attention retrieval mechanism to align fine-grained appearance features with the geometry-edited representation. This design facilitates both precise geometry editing and faithful appearance preservation in object composition.}
  \label{fig: Framework}
\end{figure*}

\section{Related Work}
The longstanding challenge in image generation is to achieve both geometric editability and visual consistency~\cite{chen2025contrastive, xu2025exploiting, Xu2021TextGuidedHI}, a problem particularly acute in object composition. This section reviews the evolution of object composition techniques, highlighting the persistent trade-off between these two competing goals.

Early approaches addressed object composition primarily as an image harmonization~\cite{guerreiro2023pct, jiang2021ssh, ke2022harmonizer, xue2022dccf} or blending~\cite{perez2023poisson, wu2019gp, zhang2021deep} task. While methods like DCCF~\cite{xue2022dccf} excel at adjusting color and lighting statistics, they operate under the assumption of fixed geometries, entirely neglecting spatial alignment. In parallel, matting-based approaches~\cite{li2022bridging} and GAN-based frameworks~\cite{azadi2020compositional, park2019gaugan, yildirim2025warping} attempted to model interactions or extract accurate foregrounds. However, they remain limited by their reliance on precise segmentation maps or struggle to handle complex out-of-plane transformations~\cite{zhan2019adaptive}, failing to generate the novel geometric structures required for realistic composition.

The advent of diffusion models~\cite{nichol2021improved, song2020denoising, ho2022classifier, li2025diffusion, xie2025mosaicfusion, feng2025diffusion} has significantly advanced the field by introducing powerful generative priors.
Initial optimization-based techniques, such as DreamBooth~\cite{ruiz2023dreambooth} and Textual Inversion~\cite{gal2022image}, offered high-fidelity customization by fine-tuning model weights or learning a unique textual token on subject-specific images. While effective in capturing visual details, these methods require time-consuming test-time optimization for each concept and lack direct control over the composed geometry, making them impractical for interactive applications.
To overcome these limitations and enable efficient feed-forward generation, encoder-based approaches have emerged. ELITE~\cite{wei2023elite} maps visual concepts into the textual embedding space for fast customization, while InstantID~\cite{wang2024instantid} and PhotoMaker~\cite{li2024photomaker} employ strong ID extractors or stacked embedding strategies to achieve zero-shot identity preservation. Nevertheless, these methods are primarily tailored for facial identity or global style transfer. They often treat the target object as a global concept rather than a spatial entity, lacking the spatial controllability required to seamlessly composite generic objects into varying background geometries.

To specifically address the challenge of generic object editability, recent research has bifurcated into two main directions, each with its own limitations.
The first direction leverages compact semantic embeddings to enable flexible manipulation. Methods such as Pbe~\cite{yang2023paint}, ObjectStitch~\cite{song2023objectstitch}, and TF-Icon~\cite{lu2023tf} project the reference image into the CLIP or DINO embedding space to guide the generation via cross-attention. This high-level semantic representation allows for significant geometric changes (e.g., pose variability); however, due to the inherent compression in CLIP, these embeddings inevitably discard instance-specific high-frequency details, resulting in a loss of texture fidelity or semantic drift.
The second direction seeks to preserve these details by introducing structural guidance or pixel-wise reference features. Approaches like AnyDoor~\cite{chen2024anydoor} and ControlNet~\cite{zhang2023adding} incorporate rigid structural representations (e.g., Canny edges or high-frequency maps) to improve fidelity. While this ensures texture sharpness, the strict structural conditioning locks the object's geometry, constraining the model's ability to adapt the object to new poses. In parallel, attention-based methods like MaskDiffusion~\cite{Zhou2023MaskDiffusionBT} have introduced adaptive masks to enhance cross-modality alignment, yet they primarily target text-to-image consistency rather than reference-based object preservation. Similarly, MimicBrush~\cite{zhao2024zero} utilizes a parallel reference U-Net to extract appearance features. Yet, without an explicit mechanism to spatially warp these features, it struggles with misalignment, often leading to ``copy-paste" artifacts rather than adaptive transformations.

Most recently, large-scale foundation model-based methods have attempted to bypass these issues by scaling up architectures and data. Insertanything~\cite{song2025insert} and UniReal~\cite{chen2025unireal} leverage the massive priors of Diffusion Transformer (DiT) architectures (e.g., Flux~\cite{Labs2025FLUX1KF}) or large-scale video datasets, utilizing powerful generative capacities to synthesize realistic lighting and physical interactions. Expanding the scope to unified frameworks, DreamOmni~\cite{Xia2024DreamOmniUI} and DreamVE~\cite{Xia2025DreamVEUI} propose to jointly train generation and editing tasks across image and video domains, while DreamOmni2~\cite{Xia2025DreamOmni2MI} further integrates Vision-Language Models (VLMs) to enable complex multimodal instruction based control. 

In summary, current research faces a bifurcated dilemma: lightweight approaches often struggle to balance geometric flexibility with appearance fidelity, whereas massive-scale models achieve quality at the expense of prohibitive computational burdens. In contrast, our DGAD framework offers an elegant solution to this editability-fidelity dilemma without resorting to massive scaling. By explicitly disentangling implicit geometry editing from appearance preservation, DGAD achieves precise structural manipulation and high-fidelity texture retention efficiently.

\section{Method}
The overall pipeline of the DGAD framework is illustrated in Fig.~\ref{fig: Framework}.  
In the encoding stage, we utilize CLIP-derived semantic embeddings and a cross-attention mechanism built upon the strong spatial perception capabilities of pretrained diffusion models to implicitly capture object geometry. During decoding, a dense cross-attention mechanism is employed to explicitly retrieve and positionally align appearance features with their corresponding geometric regions based on the encoded features.
In the following sections, we first describe the learning process of the geometry-editable representation (Section~\ref{Encoding}),  then detail the appearance-preserving representation retrieval and learning (Section~\ref{Decoding}), and finally present the optimization procedure of the proposed DGAD framework (Section~\ref{optimization}).

\subsection{Geometry-Editable Encoder}
\label{Encoding}  
This section details the learning process for capturing implicit geometric properties of objects to enhance editing capabilities. While current approaches combining semantic embeddings \cite{radford2021learning, oquab2023dinov2} with pretrained diffusion models enable basic geometric editability, existing integration strategies remain limited in both flexibility and effectiveness.  Some methods \cite{ye2023ip, yang2023paint} formulate the GOC task as image inpainting and attempt to model object geometry through end-to-end training, but their practical application is constrained by dependence on fine-grained masks.  Alternative  approaches \cite{chen2024anydoor} employ ControlNet-based architectures conditioned on specified regions but fall short in fully capturing geometric structure due to architectural limitations. Inspired by recent work \cite{su2025chord} that demonstrates introducing an intermediate 2D layout representation greatly strengthens the geometric perception of the generative model, we propose directly concatenating user-specified regions with surrounding contextual information as initial input, thereby utilizing the pretrained diffusion model’s geometric perception ability to simplify the learning process.

Given paired training data $(\mathbf{I}_o, \mathbf{I}_b, \mathbf{M}, \mathbf{I}_t)$, our process begins by encoding the object $\mathbf{I}_o$ into a compact semantic vector $\mathbf{v}_s$ using a pre-trained CLIP/DINO encoder. Concurrently, the background $\mathbf{I}_b$ and target $\mathbf{I}_t$ are encoded into latent representations, $\mathbf{z}_b$ and $\mathbf{z}_0$, via the VAE of the pretrained diffusion model. The mask $\mathbf{M}$ is downsampled to match the latent space resolution, and the noisy latent $\mathbf{z}_t$ is obtained by corrupting $\mathbf{z}_0$ with noise $\epsilon$ at timestep $t$. With these components prepared, we construct a multi-channel input by concatenating the downsampled mask $\mathbf{M}$, the background latent $\mathbf{z}_b$, and the noisy target $\mathbf{z}_t$. Here, $\mathbf{M}$ acts as a robust spatial prior, triggering the pre-trained diffusion model's inherent capacity for geometric perception. To fully harness this prior without disrupting the feature distribution, we initialize the weights for these expanded channels by replicating the model's original input weights.

Building on this spatial foundation, the cross-attention mechanism functions as the engine for \textit{geometry synthesis}. Crucially, since the semantic vector $\mathbf{v}_s$ is spatially invariant (containing only identity cues), it cannot dictate the object's pose or shape. Instead, the burden of geometric structuring falls on the backbone feature $\mathbf{h}$, which acts as the query. Driven by the mask prior, $\mathbf{h}$ proactively retrieves and reshapes the semantic information from $\mathbf{v}_s$ (the key/value) to fit the harmonious geometry inferred from the background context:
\begin{equation}
    \mathbf{Q} = \mathbf{h} \mathbf{W}_Q, \quad \mathbf{K} = \mathbf{v}_s \mathbf{W}_K, \quad \mathbf{V} = \mathbf{v}_s \mathbf{W}_V
\end{equation}
\begin{equation}
    \text{Attention}(\mathbf{Q}, \mathbf{K}, \mathbf{V}) = \text{Softmax}\left(\frac{\mathbf{Q}\mathbf{K}^\top}{\sqrt{d}}\right) \mathbf{V}
\end{equation}
where $\mathbf{W}_Q, \mathbf{W}_K, \mathbf{W}_V$ are learnable projection matrices. By decoupling semantic identity from spatial structure, this interaction forces the model to generate a geometry-editable representation that aligns with the background. Finally, the model is optimized via:
\begin{equation}
    \mathcal{L} = \mathbb{E}_{\mathbf{z}_0, \epsilon \sim \mathcal{N}(0, \mathbf{I}), t} \left[ \left\| \epsilon - \epsilon_\theta \left( \mathbf{z}_t, t, \mathbf{v}_s, \mathbf{z}_b, \mathbf{M} \right) \right\|^2 \right]
\end{equation}
where the conditioning ensures the trained model can flexibly manipulate object geometry from simple region specifications.

\subsection{Appearance-Preserving Decoder}
\label{Decoding}
The learned geometry-editable representation fails to preserve object appearance fidelity, primarily because semantic vectors predominantly encode high-level semantic cues while neglecting fine-grained visual details. Recent advances~\cite{ju2024brushnet, hu2024animate} have demonstrated that appearance features extracted from reference networks can effectively reconstruct objects from noisy latents—for instance, BrushNet~\cite{ju2024brushnet} trained on open-domain data achieves direct object reconstruction using only a single reference image. Building on these insights, we introduce a dense cross-attention mechanism that explicitly retrieves and positionally aligns appearance features with their corresponding geometric regions based on the geometry-edited representation, thereby ensuring both precise geometry editing and faithful appearance preservation.

Specifically, let \(\mathbf{h} \in \mathbb{R}^{B \times C \times H \times W}\) denote the geometry-editable backbone feature from the encoder, and \(\mathbf{F}_r \in \mathbb{R}^{B \times C \times H \times W}\) represent the fine-grained appearance feature extracted by the reference network. To spatially map the appearance details onto the learned structure (as visualized in Fig.~\ref{fig:visual_ablation_encoder}), we designate the reshaped \(\mathbf{h}\) as the query (\(\mathbf{Q}\)) and \(\mathbf{F}_r\) as the key (\(\mathbf{K}\)) and value (\(\mathbf{V}\)):
\begin{equation}
    \mathbf{Q} = \mathbf{h} \mathbf{W}_Q, \quad \mathbf{K} = \mathbf{F}_r \mathbf{W}_K, \quad \mathbf{V} = \mathbf{F}_r \mathbf{W}_V 
\end{equation}
Instead of relying on standard cross-attention for implicit retrieval, we introduce a \textbf{dense attention mechanism} that explicitly gates the flow of appearance information based on the learned geometry. At its core, this mechanism learns a position-wise softmask, \(\mathbf{m}\), directly from the query \(\mathbf{Q}\), which spatially identifies the object's regions.
\begin{equation}
    \mathbf{m} = \sigma(\mathcal{F}(\mathbf{Q})) \in [0, 1]^{B \times 1 \times H \times W}
\end{equation}
where \(\mathcal{F}\) is a lightweight MLP with a sigmoid activation \(\sigma\). This mask enables a spatially controlled fusion of appearance and geometry. We define the fused output \(\mathbf{h}_{out}\) as:
\begin{equation} 
\begin{split}
    \mathbf{h}_{out} = &\left( \text{Softmax}\left( \frac{\mathbf{Q}\mathbf{K}^\top}{\sqrt{d}} \right) \mathbf{V} \right) \odot \mathbf{m} + \mathbf{h} \odot (1 - \mathbf{m})
\end{split}
\label{eq:fusion}
\end{equation}
This formulation has a dual effect: (1) The \(\odot \mathbf{m}\) term ensures that retrieved appearance features are applied \textit{only} within the object's geometric boundaries. (2) The \(\odot (1 - \mathbf{m})\) term acts as a residual connection, preserving the original backbone features \(\mathbf{h}\) (containing scene context from the encoder) in the background. The resulting \(\mathbf{h}_{out}\), now enriched with fine-grained textures, propagates through the subsequent decoder layers. This explicit gating mechanism establishes a dense correspondence between \(\mathbf{h}\) and \(\mathbf{F}_r\), ensuring high-fidelity appearance preservation.

\subsection{Optimization}
\label{optimization}
In our primary implementation, DGAD is trained end-to-end based on the pretrained Stable Diffusion v1.5 model. Notably, our framework is model-agnostic and can be readily adapted to advanced backbones such as SDXL, as demonstrated in our experiments. A semantic-guided cross-attention mechanism is applied to every block in the backbone to learn the geometry-editable representation. In contrast, the dense attention mechanism is applied only in the decoder stage to explicitly retrieve and spatially align appearance-preserving features with their corresponding geometric regions, conditioned on the fully learned editable representation. The final loss is defined as follows:
\begin{equation}
    \mathcal{L} = \mathbb{E}_{\mathbf{z}_0, \epsilon \sim \mathcal{N}(0, \mathbf{I}), t} \left[ \left\| \epsilon - \epsilon_\theta \left( \mathbf{z}_t, t, \mathbf{v}_s, \mathbf{z}_b, \mathbf{M}, \mathbf{F}_r \right) \right\|^2 \right]
    \label{eq:loss}
\end{equation}
where the network $\epsilon_\theta$ is conditioned on the semantic vector $\mathbf{v}_s$, background latent $\mathbf{z}_b$, mask $\mathbf{M}$, and reference features $\mathbf{F}_r$. This optimization updates learnable parameters via diffusion-constrained backpropagation, enabling simultaneous precision in geometry manipulation and visual consistency maintenance during object composition.

\begin{table*}[!t]
\centering
\caption{Competing with existing baselines. Metrics are grouped by functionality: editability, appearance preservation, semantic consistency, and inference efficiency. Our method achieves superior performance across all categories while maintaining competitive efficiency. All inference times are measured on a single NVIDIA RTX 3090 GPU.}
\resizebox{\textwidth}{!}{
\begin{tabular}{l|cc|cc|cc|cc} 
\toprule
Metrics
& \multicolumn{2}{c|}{\textbf{Editability}} 
& \multicolumn{2}{c|}{\textbf{Appearance Preservation}} 
& \multicolumn{2}{c|}{\textbf{Semantic Consistency}} 
& \multicolumn{2}{c}{\textbf{Efficiency}} \\ 
\midrule
Methods
& IR $\uparrow$ & FID $\downarrow$ 
& LPIPS $\downarrow$ & DISTS $\downarrow$ 
& CLIP Score $\uparrow$ & DINO Score $\uparrow$ 
& Params (B) & Time (s) \\ 
\midrule
Ipadapter     & 42.56 & 31.35 & 18.47 & 23.46 & 87.78 & 65.09 & 1.32 & 5.80 \\ 
Objectstitch  & 41.21 & 32.25 & 18.96 & 23.53 & 87.06 & 65.09 & 1.31 & 6.04 \\
Pbe           & 43.25 & 32.07 & 20.49 & 26.48 & 85.30 & 62.14 & 1.34 & 5.24 \\
Mimicbrush    & 44.88 & 30.69 & 15.33 & 19.20 & 89.34 & 69.98 & 2.53 & 7.73 \\
Anydoor       & 44.81 & 26.08 & 15.82 & 19.21 & 88.21 & 69.22 & 2.45 & 9.42 \\
\textbf{Ours} & \textbf{61.14} & \textbf{15.04} & \textbf{14.94} & \textbf{18.53} & \textbf{89.38} & \textbf{69.92} & 2.71 & 8.34 \\
\bottomrule
\end{tabular}
}
\label{table:comparison_general}
\end{table*}

\begin{table*}[t]
    \centering
    \caption{Comparison between InsertAnything (Flux-based) and our DGAD (SDXL-based). Our method achieves superior realism (FID) and efficiency with only minor trade-offs in texture metrics. \textcolor{betterred}{Red} indicates improvement, while \textcolor{worsegreen}{Green} indicates degradation. The comparison column shows the absolute difference ($\Delta$) or relative factor.}
    \renewcommand{\arraystretch}{1.2}
    \setlength{\tabcolsep}{10pt} 

    \begin{tabular}{l l c c c}
        \toprule
        \multirow{2}{*}{\textbf{Category}} & 
        \multirow{2}{*}{\textbf{Metric}} & 
        \textbf{InsertAnything} & \textbf{Ours} & 
        \multirow{2}{*}{\textbf{Comparison ($\Delta$)}} \\
         & & (Flux) & (SDXL) & \\
        \midrule

        \multirow{4}{*}{Quality}
          & ImageReward $\uparrow$ & \textbf{69.37} & 67.96 & \loss{1.41} \\ 
          & FID $\downarrow$        & 18.54 & \textbf{13.57} & \gain{4.97} \\ 
          & LPIPS $\downarrow$      & \textbf{12.13} & 13.24 & \loss{1.11} \\ 
          & DISTS $\downarrow$      & \textbf{15.49} & 16.02 & \loss{0.53} \\ 
        \midrule

        \multirow{2}{*}{Semantic}
          & CLIP Score $\uparrow$   & \textbf{92.08} & 90.15 & \loss{1.93} \\ 
          & DINO Score $\uparrow$   & \textbf{73.99} & 71.34 & \loss{2.65} \\ 
        \midrule

        \multirow{2}{*}{\textbf{Efficiency}}
          & Params (B) $\downarrow$ & 18.29 & \textbf{4.90} & \gain{3.7$\times$ Smaller} \\
          & Time (s) $\downarrow$   & 36.72 & \textbf{10.53} & \gain{3.5$\times$ Faster} \\
        \bottomrule
    \end{tabular}

    \label{table:comparison_sdxl}
\end{table*}

\section{Experiments}
\subsection{Evaluation Benchmark}
\noindent\textbf{Competing Algorithms.} 
We compare our method with several recent approaches in the field of object composition. Since the majority of representative baselines (e.g., AnyDoor \cite{chen2024anydoor}, PbE \cite{yang2023paint}) are built upon Stable Diffusion v1.5~\cite{Rombach2021HighResolutionIS}, we implement our primary DGAD model on this backbone to ensure a fair and direct comparison. 
Furthermore, acknowledging the recent trend of introducing larger architectures, such as InsertAnything~\cite{song2025insert} (based on Flux \cite{Labs2025FLUX1KF}), we also seek to validate the effectiveness of our method across different backbones. To this end, while balancing performance with inference efficiency, we implement a variant of DGAD using SDXL \cite{Podell2023SDXLIL} as the backbone and compare it against these large-scale foundation models \cite{song2025insert}.
Specifically, the competing methods include:
\begin{enumerate}[label=\arabic*)]
    \item \textbf{IP-Adapter}~\cite{ye2023ip}: Formulates object composition as an inpainting task, using CLIP-derived semantic features and end-to-end fine-tuning for generation.
    \item \textbf{PbE}~\cite{yang2023paint} (CVPR 2023): Employs CLIP-derived semantic features and strong data augmentations to achieve object composition. 
    \item \textbf{ObjectStitch}~\cite{song2023objectstitch} (CVPR 2023): Introduces a content adaptor to preserve both categorical semantics and object appearance during composition. 
    \item \textbf{AnyDoor}~\cite{chen2024anydoor} (CVPR 2024): Utilizes semantic features extracted from DINO and structured representation via ControlNet to perform object composition. 
    \item \textbf{MimicBrush}~\cite{zhao2024zero} (NeurIPS 2024): Leverages appearance features from a reference network to guide object composition. 
    \item \textbf{InsertAnything}~\cite{song2025insert} (AAAI 2026): A recent approach that builds upon the large-scale Flux architecture to exploit powerful generative priors.
\end{enumerate}

\noindent\textbf{Dataset.} Following the approach in AnyDoor \cite{chen2024anydoor}, we use a training set consisting of 386k images and 23k video samples. For the image dataset, we apply LaMa \cite{suvorov2022resolution} to remove foreground objects, creating paired data that includes the object, the background scene, and the corresponding target image. For the video dataset, we also adopt AnyDoor’s preprocessing method to construct triplets with the same structure as the image data. For evaluation, we select 30 object concepts from DreamBooth \cite{ruiz2023dreambooth} as test subjects. As background scenes, we manually choose 80 geometrically annotated images from the COCO-Val set \cite{lin2014microsoft}, resulting in 2,400 synthesized samples covering all combinations of the selected objects and scenes.
Unless otherwise specified, all experiments are conducted at 512×512 resolution.

\noindent\textbf{Metrics.} 
The purpose of object composition is to edit objects with desired geometric properties to align with the background scene while preserving their appearance details. 
\textbf{To evaluate the object Editability}, we introduce two metrics: IR \cite{Xu2023ImageRewardLA}, and FID \cite{Heusel2017GANsTB}. IR is text-to-image evaluation models trained on large-scale datasets that reflect human preferences for generated images. Since high-quality object edits tend to produce more visually appealing results that align with human expectations, it serve as human-aligned indicators of editing success. FID assesses the distributional similarity between the composed images and real-world images, providing an objective measure of compositional realism. 
\textbf{To evaluate appearance consistency}, we introduce two metrics: LPIPS \cite{Zhang2018TheUE} and DISTS \cite{Ding2020ImageQA}. These metrics can robustly assess an object's appearance consistency even under geometric misalignment by measuring differences in deep feature space.
Additionally, we introduce CLIP Score \cite{Hessel2021CLIPScoreAR} and DINO Score \cite{oquab2023dinov2} \textbf{to measure the semantic consistency} between objects as a supplement, similar to \cite{chen2024anydoor}.

\subsection{Quantitative Comparison}
In this section, we present a comprehensive quantitative evaluation of DGAD. To ensure a rigorous and fair assessment, we structure our analysis into two distinct perspectives: 
(1) Competing with existing baselines: Since the majority of prior arts utilize the Stable Diffusion v1.5~\cite{Rombach2021HighResolutionIS} backbone, we benchmark our primary model against them on the same architecture to ensure a fair comparison (Table \ref{table:comparison_general}).
(2) Generalization to large-scale foundation models: We extend DGAD to the SDXL \cite{Podell2023SDXLIL} backbone to verify its scalability. This analysis serves to demonstrate that our framework is model-agnostic and capable of leveraging stronger generative priors from advanced architectures. Crucially, it offers a pragmatic alternative that balances high-quality generation with superior inference efficiency compared to computationally intensive massive models \cite{song2025insert} (Table \ref{table:comparison_sdxl}).

\vspace{1mm}
\noindent\textbf{Competing with existing baselines.}
As presented in Table \ref{table:comparison_general}, we first compare DGAD with representative methods utilizing comparable backbone architectures.
IPAdapter \cite{ye2023ip}, ObjectStitch \cite{song2023objectstitch}, and Pbe \cite{yang2023paint} benefit from lightweight architectures ($\sim$1.3B parameters) and fast inference speeds ($\sim$5-6s). However, relying solely on CLIP embeddings results in a significant loss of high-frequency details, evidenced by their inferior appearance preservation scores (e.g., LPIPS $>$ 18.0). Furthermore, their inability to seamlessly integrate realistic objects leads to weaker editability, with FID scores hovering above 31.0. AnyDoor \cite{chen2024anydoor} introduces structured representations to improve fidelity. While this lowers LPIPS to 15.82, the rigid structural conditioning restricts geometric flexibility (IR 44.81). Notably, despite having fewer parameters (2.45B) than our model, AnyDoor exhibits the highest inference latency (9.42s). 
MimicBrush \cite{zhao2024zero} achieves competitive appearance preservation (LPIPS 15.33) but struggles with explicit correspondence, limiting its editability (IR 44.88). 

In contrast, our DGAD achieves the best performance across all quality metrics in this group. By leveraging the inherent spatial perception of diffusion models, we boost the editability score significantly to 61.14 (IR) and reduce the FID to 15.04. Crucially, although incorporating a reference network increases our parameter count to 2.71B, our design ensures highly efficient inference (8.34s), outperforming AnyDoor in both speed and quality.

\vspace{1mm}
\noindent\textbf{Generalization to large-scale foundation models.} 
To demonstrate the generalizability of our framework, while strategically prioritizing the balance between generation quality and computational cost, we extended DGAD to the SDXL backbone. We compare this efficient variant against the state-of-the-art Flux-based method, InsertAnything~\cite{song2025insert}. 
As shown in Table~\ref{table:comparison_sdxl}, InsertAnything benefits from the massive capacity of the Flux~\cite{Labs2025FLUX1KF} architecture (18.29B parameters), achieving slightly better texture fidelity (e.g., lower LPIPS). 
However, this marginal gain comes with significant computational costs. 
In contrast, our DGAD-SDXL achieves a significantly superior FID (13.57 vs. 18.54), improving the metric by nearly 5 points. 
Given that FID measures the distributional divergence from real images, this result confirms that while Flux-based methods focus on rigid detail retention, our disentangled approach strikes a better balance between identity preservation and scene-level harmonization, generating composites that align more naturally with the real-world distribution.
Crucially, DGAD demonstrates a decisive advantage in efficiency. By avoiding the heavy computational burden of 18B parameters, our method reduces the model size by 3.7$\times$ and achieves $\sim$3.5$\times$ faster inference speed (10.53s vs. 36.72s). This positions DGAD as a highly practical solution for real-world applications where interactive latency is a critical constraint.

\subsection{Qualitative Comparisons}

\begin{figure*}[!t]
    \centering
    \includegraphics[width=0.98\textwidth]{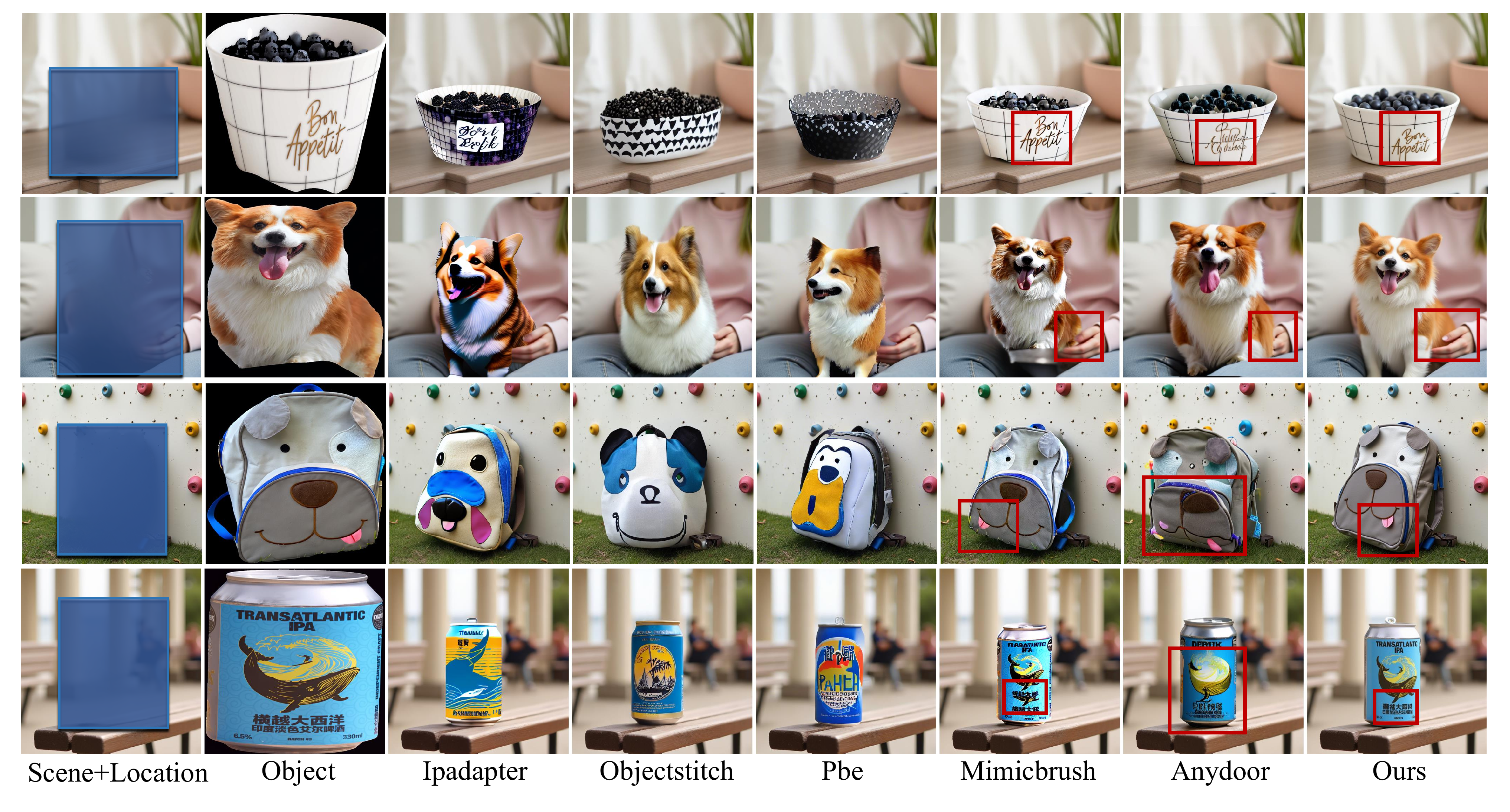}
    \vspace{-10pt}
    \caption{Qualitative comparison under coarse bounding-box guidance (Implicit Geometry). Inputs are defined only by rough boxes (blue). While competitors often struggle to hallucinate plausible geometries or preserve identity, DGAD implicitly learns natural object poses (e.g., the sitting dog, the bag's deformation) while retaining fine-grained appearance.}
    \label{fig:visual_baseline}
\end{figure*}

\begin{figure*}[!t]
    \centering
    \includegraphics[width=0.98\textwidth]{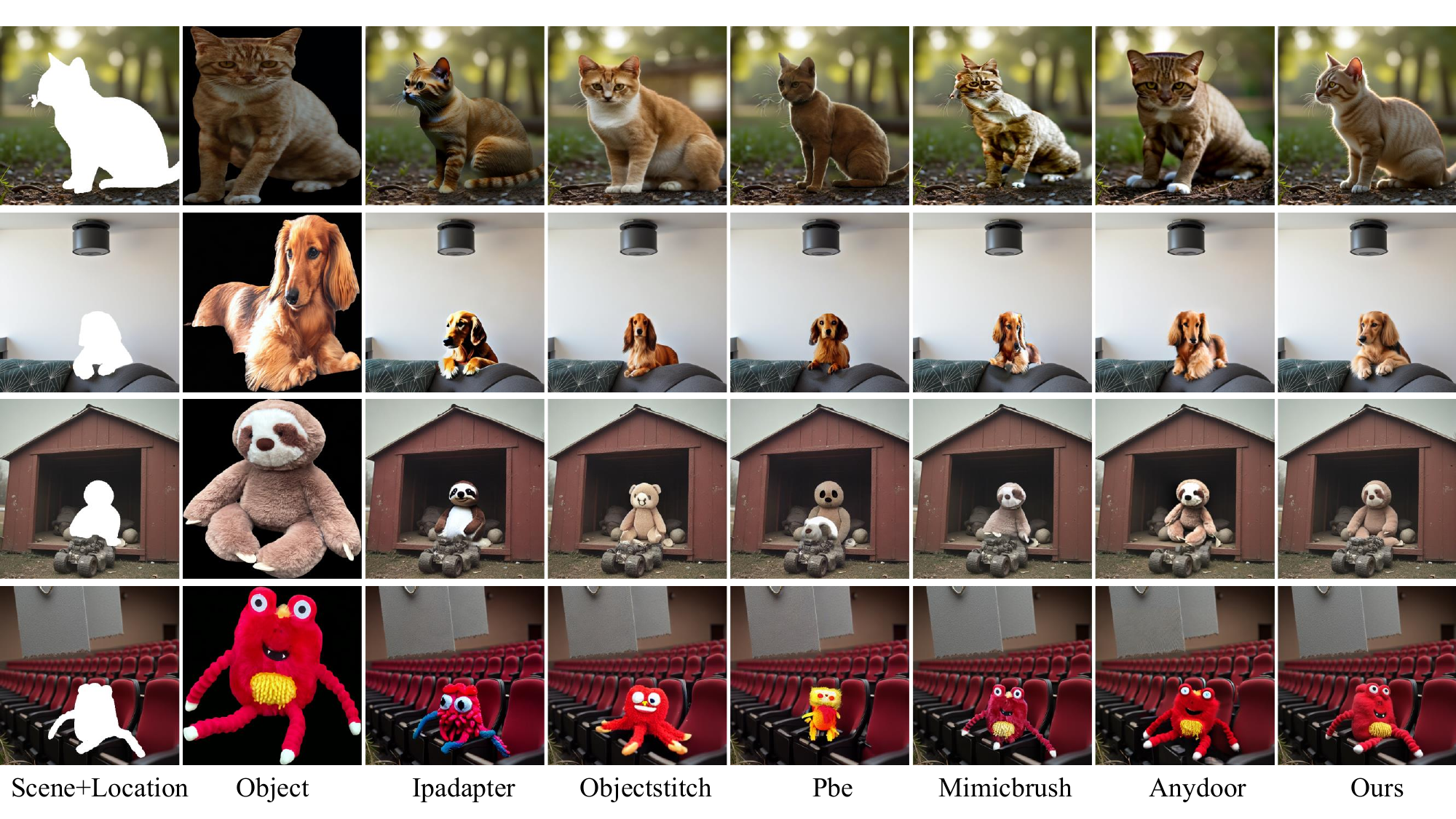}
    \vspace{-10pt}
    \caption{Qualitative comparison under fine-grained mask guidance (Explicit Shape Control). Inputs are defined by specific silhouettes (white). Semantic-driven methods (e.g., PbE) suffer from semantic drift (Row 3), and reference-based methods (e.g., MimicBrush) struggle with texture warping. DGAD faithfully fills the specified shape with high-fidelity textures.}
    \label{fig:visual_baseline_shape}
\end{figure*}

In this section, we present a comprehensive qualitative comparison between DGAD and representative methods. Consistent with our primary comparative settings, the visual results shown in Fig.~\ref{fig:visual_baseline} and Fig.~\ref{fig:visual_baseline_shape} are obtained using the Stable Diffusion v1.5 backbone. We provide a detailed analysis of the performance differences relative to the architectural mechanisms of each method. Note that for visual comparisons regarding the scalability analysis (i.e., DGAD-SDXL vs. InsertAnything-Flux), please refer to the Supplementary Material.

\noindent\textbf{Generation from Coarse Bounding Boxes (Fig.~\ref{fig:visual_baseline}).} 
We first evaluate the capability to implicitly learn geometry from rough bounding box inputs. 
IPAdapter, ObjectStitch, and PbE rely on CLIP-derived semantic embeddings. Since CLIP primarily captures category-level semantics, these methods inevitably lose fine-grained details. As shown in Fig.~\ref{fig:visual_baseline} (Cols. 3-5), while they can generate geometrically plausible objects, they fail to preserve the specific visual identity (e.g., the texture of the backpack in Fig.~\ref{fig:visual_baseline}, Row 3).
AnyDoor utilizes DINO features and Canny edges. However, its reliance on rigid structural constraints limits flexibility; it tends to simply ``copy-paste'' the reference image with minimal geometric adaptation, leading to unnatural transitions (e.g., the dog in Fig.~\ref{fig:visual_baseline}, Row 2 does not naturally sit on the couch).
MimicBrush achieves better texture retention but lacks explicit spatial correspondence. Under complex geometric transformations, it often suffers from severe structure-texture mismatches (e.g., the distorted features of the bag in Fig.~\ref{fig:visual_baseline}, Row 3).
In contrast, DGAD successfully leverages semantic priors to synthesize natural geometries (e.g., a dog sitting comfortably) while using dense attention to rigorously map fine-grained textures, achieving the best perceptual quality.

\vspace{1mm}
\noindent\textbf{Versatility with Fine-Grained Mask Guidance (Fig.~\ref{fig:visual_baseline_shape}).}
Real-world editing often demands precise shape control. Fig.~\ref{fig:visual_baseline_shape} demonstrates the performance when strictly constrained by user-provided masks.
Semantic-driven methods (PbE, ObjectStitch) are prone to semantic drift when the target shape deviates from the reference's canonical pose. A striking example is visible in Fig.~\ref{fig:visual_baseline_shape} (Row 3): PbE misinterprets the sitting sloth shape, generating a bear-like figure instead.
Reference-based methods (MimicBrush, AnyDoor) struggle to warp intricate textures into the target shape without artifacts. For instance, in Fig.~\ref{fig:visual_baseline_shape} (Row 4) (the red monster), MimicBrush fails to align the complex facial features with the new pose, resulting in blurred textures.
DGAD exhibits superior versatility. Thanks to our disentangled design, the encoder respects the structural constraint of the mask, while the decoder's dense cross-attention ensures that the specific identity (e.g., the monster's fur and eyes) is meticulously reconstructed within the user-defined boundary. This confirms that DGAD offers both high-level flexibility and precise controllability.

\subsection{User Study}
To comprehensively evaluate the perceptual performance of our method against representative methods based on Stable Diffusion v1.5, we conducted a rigorous user study. A total of 25 participants were invited to evaluate 30 sets of randomly sampled test cases. Each set consisted of a background scene with a specified mask and results generated by six different methods. Participants assessed the results based on two distinct criteria: \textbf{Composition Quality}, which measures the geometric alignment and naturalness of the inserted object within the scene context, and \textbf{Visual Consistency}, which evaluates the fidelity of texture, color, and material preservation relative to the reference object.

\begin{figure}[t]
    \centering
    \includegraphics[width=\linewidth]{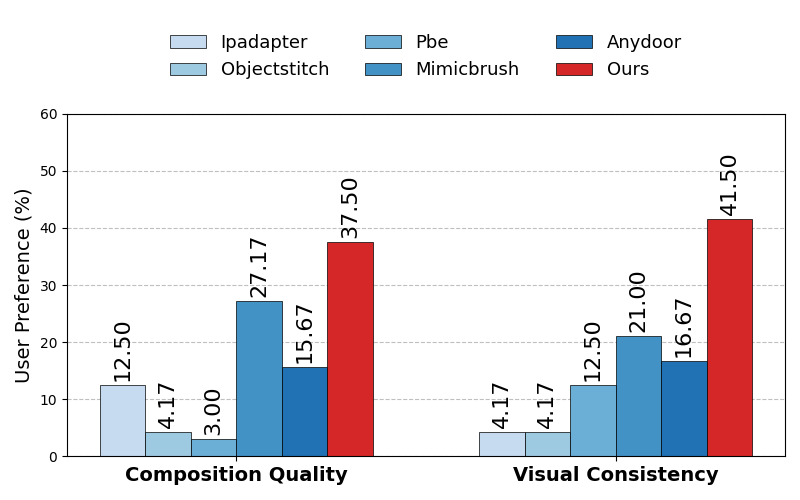}
    \caption{\textbf{User study results.} Comparison of user preferences on Composition Quality and Visual Consistency. Our method (shown in red) significantly outperforms existing approaches in both metrics.}
    \label{fig:user_study}
\end{figure}

The quantitative results are summarized in Fig.~\ref{fig:user_study}. Our proposed method demonstrates superior performance across both metrics. Specifically, in terms of Composition Quality, our method achieved the highest preference rate of 37.50\%, surpassing the second-best approach (Mimicbrush) by 8.33\%. More notably, in Visual Consistency, our method secured a dominant preference rate of 41.50\%, nearly doubling the score of the nearest competitor (21.00\%). These statistics strongly validate that our framework not only accurately aligns objects with the underlying scene geometry but also excels in preserving fine-grained appearance details, aligning well with human perceptual preferences.

\subsection{Ablation Study}

To comprehensively evaluate the effectiveness of our proposed DGAD framework, we conduct a series of ablation studies. 
Consistent with our primary comparative benchmarks, all experiments in this section are performed using the Stable Diffusion v1.5 backbone.
We begin by validating the necessity of our core disentangled design, followed by detailed analyses of the specific components within our geometry-editable encoder and appearance-preserving decoder. All results are summarized in Table~\ref{table:ablation_unified}.

\begin{table*}[t]
    \centering
    \caption{Comprehensive ablation studies of the DGAD framework. We analyze the core components, the encoder design, and the decoder design. Removing any key component or strategy leads to a notable performance degradation, demonstrating the effectiveness of our integrated design. Best results are highlighted.}
    \label{table:ablation_unified}
    \resizebox{\textwidth}{!}{
    \begin{tabular}{lcccccc}
        \toprule
        \textbf{Method} & \textbf{IR $\uparrow$} & \textbf{FID $\downarrow$} & \textbf{LPIPS $\downarrow$} & \textbf{DISTS $\downarrow$} & \textbf{CLIP Score $\uparrow$} & \textbf{DINO Score $\uparrow$} \\
        \midrule
        \multicolumn{7}{l}{\textit{\textbf{(1) Analysis of Core Framework Components}}} \\
        Ours (w/o Semantic Guidance) & 35.16 & 31.55 & 15.10 & 18.98 & 86.50 & 67.23 \\
        Ours (w/o Appearance Features) & 55.24 & 21.37 & 28.15 & 33.40 & 89.10 & 69.55 \\
        \midrule
        \multicolumn{7}{l}{\textit{\textbf{(2) Analysis of Geometry-Editable Encoder Design}}} \\
        Ours (w/o Layout Representation) & 52.82 & 26.08 & 16.02 & 19.97 & 87.12 & 68.02 \\
        Ours (w/o Copied Weights) & 58.12 & 24.67 & 15.13 & 19.32 & 88.12 & 68.18 \\
        \midrule
        \multicolumn{7}{l}{\textit{\textbf{(3) Analysis of Appearance-Preserving Decoder Design}}} \\
        Ours (w/o Dense Attention) & 58.12 & 19.18 & 16.92 & 21.93 & 88.23 & 69.12 \\
        Ours (Dense Attention on Both Stages) & 60.02 & 16.13 & 15.23 & 19.21 & 89.13 & 69.12 \\
        \midrule
        \textbf{Ours (Full Model)} & \textbf{61.14} & \textbf{15.04} & \textbf{14.94} & \textbf{18.53} & \textbf{89.38} & \textbf{69.92} \\
        \bottomrule
    \end{tabular}
    }
\end{table*}

\begin{figure}[t]
    \centering
    \includegraphics[width=\linewidth]{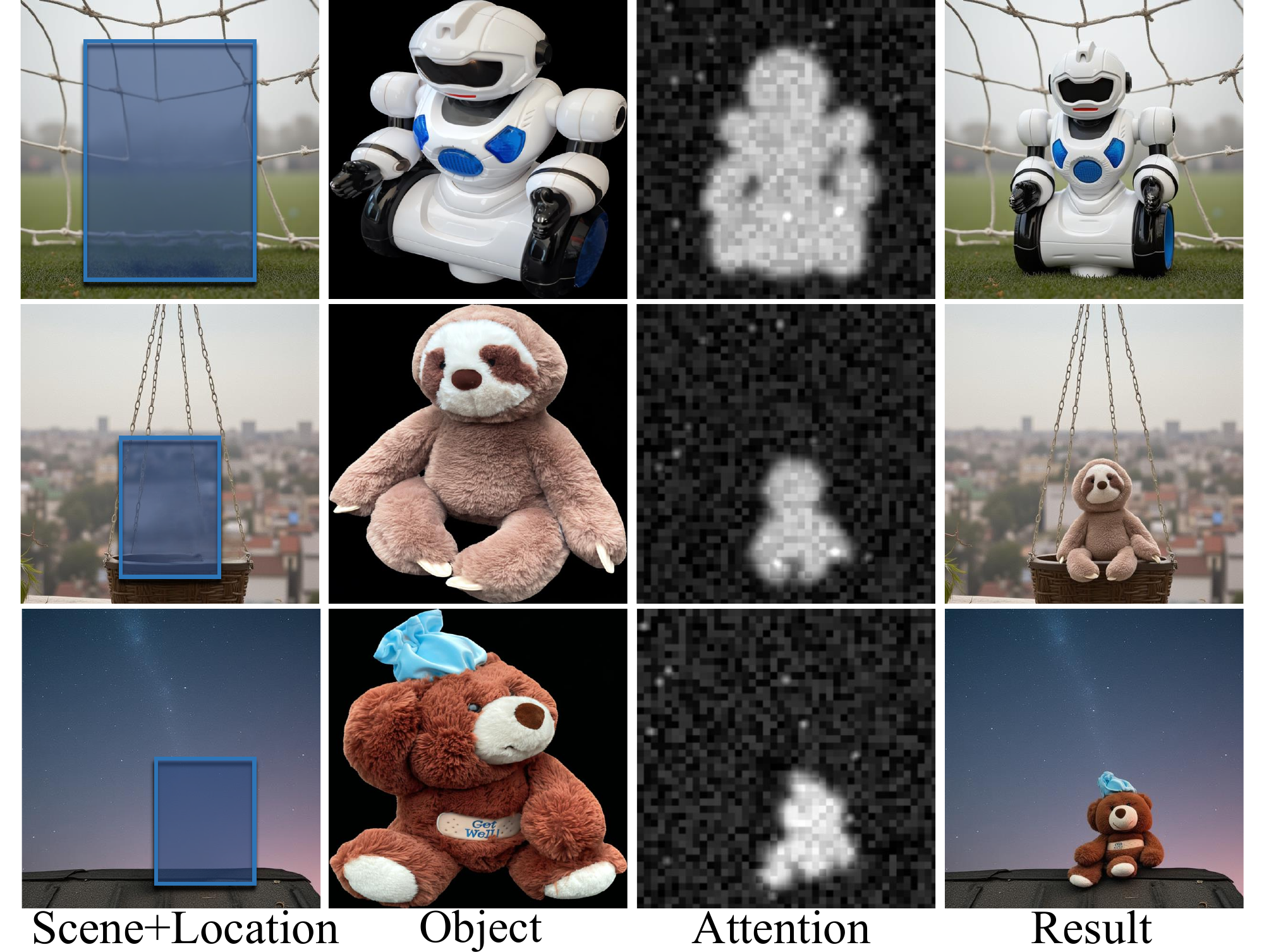}
    \caption{Visualization of the learned geometric representation. The third column shows the spatial attention maps learned by the encoder. Guided by this geometry, the model performs position-wise appearance retrieval to ensure precise texture reconstruction.}
    \label{fig:visual_ablation_encoder}
\end{figure}

\subsubsection{Effectiveness of the Disentangled Framework}
We first validate the fundamental motivation behind our framework by analyzing the contributions of its two pillars: semantic-guided geometry control and appearance feature preservation. As shown in Table~\ref{table:ablation_unified} (rows 1-2), we evaluate two variants: 

\noindent 1) \textbf{Ours (w/o Semantic Guidance)}: 
This component is designed to leverage semantic cues to elicit the pre-trained diffusion model's inherent capacity for context-aware geometry construction.
By removing the semantic embedding and relying solely on the reference network, the model retains high texture fidelity (LPIPS of 15.10), comparable to the full model. However, its geometric editability is severely compromised, with the IR score dropping to 35.16. This confirms that reference features, which are typically spatially aligned, tend to constrain the generation to the original object pose, thereby inhibiting flexible geometric manipulation.

\noindent 2) \textbf{Ours (w/o Appearance Features)}: 
This module is dedicated to retrieving and mapping fine-grained visual details onto the edited geometry.
Conversely, removing this explicit appearance injection and relying exclusively on semantic guidance yields strong editability (IR 55.24), second only to our full model. Yet, the appearance preservation degrades significantly (LPIPS worsens to 28.15). Without this mechanism, the model hallucinates a \textit{generic} instance of the object class rather than preserving the specific visual identity of the reference.

\noindent\textbf{Synergy.} These results empirically verify the central conflict in GOC: semantic embeddings facilitate shape changes but lose details, while reference features keep details but constrain the shape. Our full model effectively bridges this gap, achieving the best performance across all metrics by utilizing semantic cues for structural planning and dense attention for texture mapping.

\subsubsection{Analysis of the Geometry-Editable Encoder}

The primary function of our encoder is to synthesize a plausible geometric structure (e.g., pose and shape) from abstract semantic cues. In this section, we analyze how our specific design choices collaborate to activate and sustain this implicit geometry learning process.

We first validate that introducing a 2D layout representation (concatenated mask) as a spatial anchor effectively enables the model to perceive and construct geometry. As shown in Table~\ref{table:ablation_unified}, removing this component (``Ours w/o Layout Representation'') causes the model to degrade to a standard inpainting setup, leading to a drastic deterioration in realism with FID worsening significantly from 15.04 to 26.08. This quantitative drop confirms that semantic embeddings alone are insufficient to constrain the generation. Qualitatively, the attention visualization in Fig.~\ref{fig:visual_ablation_encoder} serves as compelling evidence: the emergence of a clear object silhouette within the masked region confirms that our layout-guided approach effectively triggers the model's inherent capacity for precise, user-specified geometric construction.

Crucially, to fully leverage this structural guidance, the model must correctly interpret the input signals using its pre-trained knowledge. This necessitates our specific weight initialization strategy. We argue that copying pre-trained weights for the expanded input channels is not merely a technical trick, but a critical step to preserve the diffusion model's spatial priors. By comparing against a baseline with randomly initialized weights (``Ours w/o Copied Weights''), we observe a notable performance drop in Table~\ref{table:ablation_unified} (FID 24.67 vs. 15.04). This degradation occurs because random initialization introduces a distributional shift that disrupts the feature extraction of subsequent layers. By utilizing weight inheritance, we ensure a seamless alignment between the new layout condition and the pre-trained spatial knowledge, effectively avoiding the ``cold start'' problem and ensuring the faithful generation of the intended geometry.

\begin{figure}[t]
    \centering
    \includegraphics[width=\linewidth]{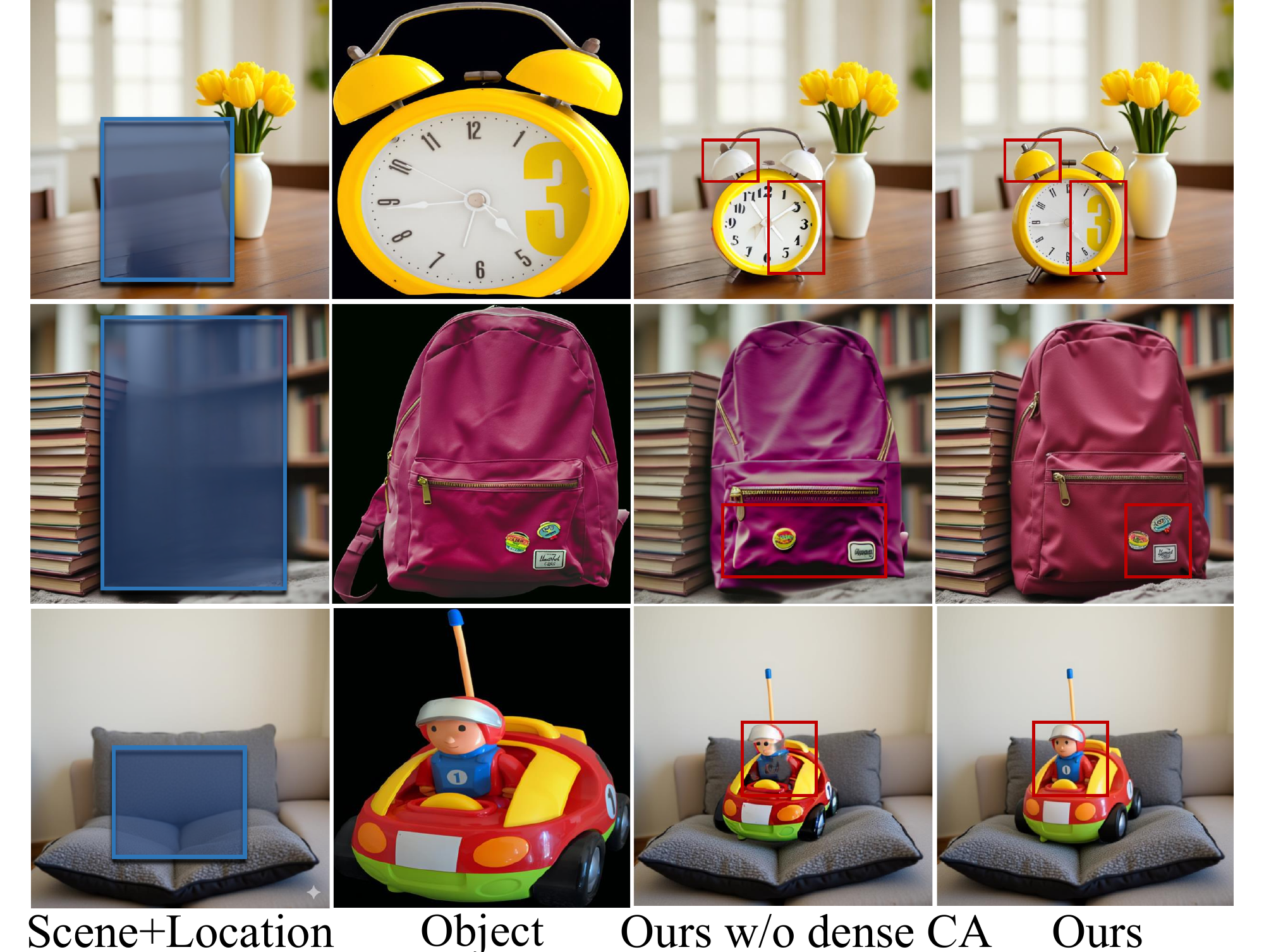}
    \vspace{-10pt}
    \caption{Visual ablation with and without Dense Cross-Attention. Unlike standard attention which loses high-frequency details, our mechanism enforces explicit feature retrieval to reconstruct sharp and accurate textures, as evident in the highlighted regions.}
    \label{fig:visual_ablation_decoder}
\end{figure}

\subsubsection{Analysis of the Appearance-Preserving Decoder}

Having successfully synthesized the geometric structure in the encoder, the decoder's primary task is to accurately retrieve and map fine-grained appearance details onto this learned geometry. In this section, we analyze the mechanisms that enable this precise texture alignment and preservation.

We first validate the efficacy of our core novelty, the Dense Cross-Attention mechanism, which is designed to enforce explicit feature alignment. As shown in Table~\ref{table:ablation_unified}, replacing our dense mechanism with standard cross-attention (``Ours w/o Dense Attention'') leads to a marked decline in fidelity, with LPIPS rising from 14.94 to 16.92. This degradation occurs because standard cross-attention performs implicit global retrieval, which tends to smear out fine-grained details across the spatial domain. In contrast, our dense mechanism enforces position-aware alignment, ensuring that texture details are accurately anchored to their corresponding geometric structures. The visual artifacts in Fig.~\ref{fig:visual_ablation_decoder} (Top) further corroborate that dense attention is essential for crisp texture restoration.

\begin{figure}[t]
\centering
\includegraphics[width=\linewidth]{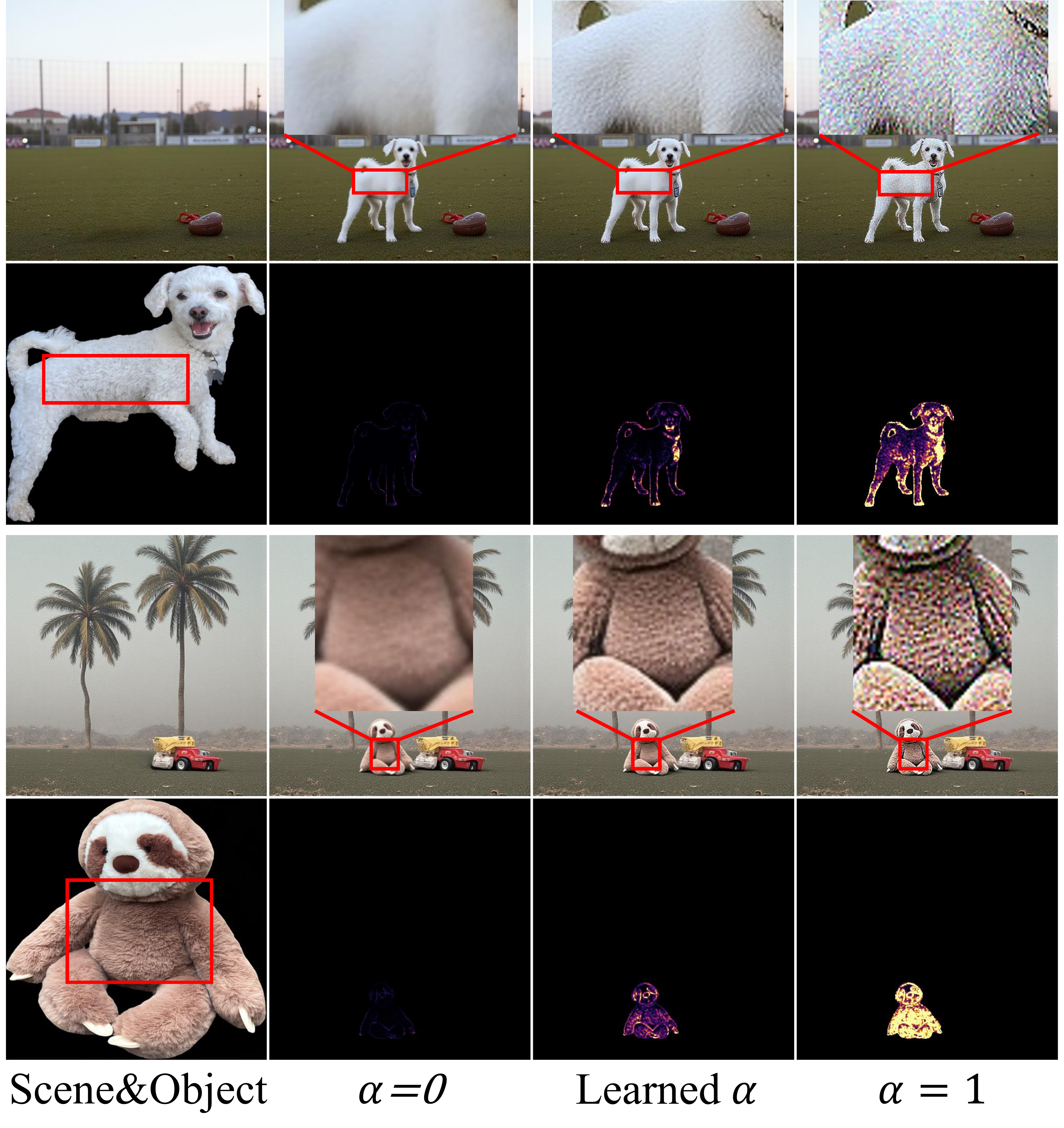}
\vspace{-10pt}
\caption{Visualization of generated results and corresponding Laplacian variance maps under different ${\alpha}$ settings.}
\label{fig:visual_ablation_decoder_Laplacian_variance}
\end{figure}

However, simply retrieving features is insufficient; the model must also intelligently regulate where and how much appearance information to inject to avoid disrupting the generated structure. To this end, we introduce a learnable mask ${\alpha}$ that functions as a spatially adaptive valve. To validate its role, we analyze the Laplacian variance maps (Fig.~\ref{fig:visual_ablation_decoder_Laplacian_variance}, Bottom) as a proxy for texture content under different settings:
\begin{itemize}
    \item \textbf{$\alpha = 0$ (Texture Poverty):} Relying solely on semantic embeddings yields geometrically correct but over-smoothed results. The suppressed response in the Laplacian maps confirms that semantic embeddings alone lack the high-frequency information needed for realistic textures.
    \item \textbf{$\alpha = 1$ (Texture Conflict):} Forcing an indiscriminate injection ($\alpha = 1$) disrupts the geometric priors established by the encoder and introduces noise into non-textured regions (e.g., background), leading to visual artifacts.
    \item \textbf{Learned $\alpha$ (Adaptive Modulation):} Our adaptive scheme selectively integrates fine-grained details only where needed. This strikes an optimal balance, preserving structural integrity while enhancing realism, as evidenced by the natural texture distribution in our Laplacian maps.
\end{itemize}

Finally, we apply the dense cross-attention mechanism exclusively in the decoder stage, based on the rationale that effective appearance retrieval should be conditioned on an already formed geometric structure. To validate this design choice, we compare our approach against a baseline where the mechanism is applied to both the encoder and decoder (``Ours (Dense Attention on Both Stages)''). Consistent with our rationale, this dual-stage application leads to a slight performance drop (Table~\ref{table:ablation_unified}). We attribute this to the functional distinction between the two stages: the encoder aims to abstract geometry from noisy latents, while the decoder aims to concretize appearance. Forcing appearance retrieval in the encoder—where features are dominated by noise and geometry is yet to be formed—introduces erroneous visual cues that conflict with structural learning. Thus, confining the mechanism to the decoder proves to be the most robust strategy for disentangled composition.

\section{Limitations and Future Work}
\label{sec:limitations}

While DGAD achieves state-of-the-art performance in our experiments, it shares inherent limitations with other 2D-based approaches. 

\noindent\textbf{Extreme View Synthesis.} 
Synthesizing views with significant occlusion (e.g., generating the back of an object) remains challenging. Since the reference view lacks texture information for these regions, the model must hallucinate content, which can lead to lower fidelity under extreme pose changes.

\noindent\textbf{Physical Consistency.} 
Without explicit 3D awareness, ensuring strict adherence to perspective laws and functional affordance in complex indoor scenes is difficult. This may occasionally result in perspective misalignment or physically inconsistent placements (e.g., floating objects).

\noindent\textbf{Future Work.} 
We aim to bridge these gaps in two directions: 
(1) Integration of 3D Priors: We plan to incorporate emerging 3D foundation models~\cite{Zhao2025Hunyuan3D2S, Lin2025Kiss3DGenRI} to infer latent 3D structures from 2D images, explicitly guiding the generation of extreme viewpoints and occluded regions.
(2) Physics and Aesthetic Reasoning: We will leverage MLLMs~\cite{Bai2023QwenTR, Achiam2023GPT4TR} as ``Scene Directors'' to analyze lighting and physical constraints, providing high-level guidance for intelligent, physics-compliant scene completion.

\section{Conclusion}
\label{sec:conclusion}
In this work, we presented DGAD, a novel framework that resolves the inherent conflict between geometric editability and appearance preservation in object composition. By explicitly disentangling these objectives, our model leverages semantic embeddings to implicitly learn geometry during encoding, while employing a dense cross-attention mechanism to spatially align fine-grained textures during decoding. This synergistic design ensures precise structural manipulation without compromising visual fidelity. Extensive experiments demonstrate that DGAD significantly outperforms state-of-the-art methods and exhibits strong scalability to large-scale foundation models like SDXL, offering a highly efficient and practical solution for high-quality generative composition.

\noindent{\textbf{Code availability.} } \quad All trained models and codes are publicly available on GitHub: \url{https://github.com/jianmanlincjx/DGAD}.

\noindent{\textbf{Data availability.} } \quad The data that support the finding of this study are openly available at the following URL: \url{https://github.com/uniBruce/Mead}, \url{https://github.com/ali-vilab/AnyDoor/blob/main/configs/datasets.yaml}.

\bibliography{sn-bibliography}
\bibliographystyle{spmpsci}      

\end{document}